\documentclass[runningheads]{llncs}
\usepackage{bbding}
\usepackage{float}

\usepackage{listings}
\usepackage{color}
\usepackage{xcolor}
\definecolor{dkgreen}{rgb}{0,0.6,0}
\definecolor{gray}{rgb}{0.5,0.5,0.5}
\definecolor{mauve}{rgb}{0.58,0,0.82}
\usepackage[T1]{fontenc}
\usepackage{graphicx}
\begin{document}

\title{Student Classroom Behavior Detection based on YOLOv7-BRA and Multi-Model Fusion}
%
%
\author{Fan Yang\inst{1} \and
Tao Wang\inst{1} \and
Xiaofei Wang\inst{2(}\Envelope\inst{)}}
\authorrunning{F. Author et al.}
%
\institute{Jinan University, Guangzhou, China\newline 
\email{winstonyf@qq.com}
\and
School of films and animation of the college of Chinese \& ASEAN arts, Chengdu University, ChengDu 610106,  China\newline
\email{wangxiaofei@cdu.edu.cn}}
\maketitle              
\begin{abstract}
Accurately detecting student behavior in classroom videos can aid in analyzing their classroom performance and improving teaching effectiveness. However, the current accuracy rate in behavior detection is low. To address this challenge, we propose the Student Classroom Behavior Detection system based on based on YOLOv7-BRA (YOLOv7 with Bi-level Routing Attention ). We identified eight different behavior patterns, including standing, sitting, speaking, listening, walking, raising hands, reading, and writing. We constructed a dataset, which contained 11,248 labels and 4,001 images, with an emphasis on the common behavior of raising hands in a classroom setting (Student Classroom Behavior dataset, SCB-Dataset). To improve detection accuracy, we added the biformer attention module to the YOLOv7 network. Finally, we fused the results from YOLOv7 CrowdHuman, SlowFast, and DeepSort models to obtain student classroom behavior data. We conducted experiments on the SCB-Dataset, and YOLOv7-BRA achieved an mAP@0.5 of 87.1\%, resulting in a 2.2\% improvement over previous results. Our SCB-dataset can be downloaded from: \url{https://github.com/Whiffe/SCB-dataset}

\end{abstract}
\section{Introduction}

\begin{figure}
\includegraphics[width=\textwidth]{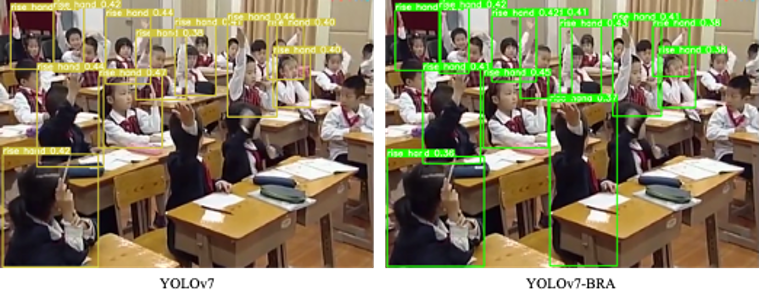}
\caption{YOLOv7 and YOLOv7-BRA detection results comparison. It is clear that YOLOv7-BRA has better detection performance.} \label{fig1}
\end{figure}

In recent years, with the development of behavior detection technology~\cite{ref_article1}, it has become possible to analyze student behavior in class videos to obtain information on their classroom status and learning performance. This technology is of great importance to teachers, administrators, students, and parents in schools. However, in traditional teaching models, teachers find it difficult to pay attention to the learning situation of every student and can only understand the effectiveness of their own teaching methods by observing a few students. School administrators rely on on-site observations and student performance reports to identify problems in education and teaching. Parents can only understand their child's learning situation through communication with teachers and students. Therefore, utilizing behavior detection technology to accurately detect student behavior and analyze their learning status and performance can provide more comprehensive and accurate feedback for education and teaching.

Existing student classroom behavior detection algorithms can be roughly divided into three categories: video-action-recognition-based~\cite{ref_article2}, pose-estimation-based~\cite{ref_article3} and object-detection-based~\cite{ref_article4}. Video-based student classroom behavior detection enables the recognition of continuous behavior, which requires labeling a large number of samples. For example, the AVA dataset~\cite{ref_article5} for SlowFast~\cite{ref_article6} detection is annotated with 1.58M. And, video behavior recognition detection is not yet mature, as in UCF101~\cite{ref_article7} and Kinetics400~\cite{ref_article8}, some actions can sometimes be determined by the context or scene alone. Pose-estimation-based algorithms characterize human behavior by obtaining position and motion information of each joint in the body, but they are not applicable for behavior detection in overcrowded classrooms. Considering the challenges at hand, object-detection-based algorithms present a promising solution. In fact, in recent years object-detection-based algorithms have made tremendous breakthroughs, such as YOLOv7~\cite{ref_article9}. Therefore, we have employed an object-detection-based algorithm in this paper to analyze student behavior.

As for object detection, the two-stage and one-stage object detection frameworks~\cite{ref_article10,ref_article11} have received more attention due to their impressive detection results on public datasets. However, the datasets from real classrooms are quite different from public ones and the classical methods perform poorly in real classrooms. One of the representative issues is large scale variations among different positions, such as students in the front row of the classroom (about 40×40 pixels) and students in the back row (about 200×200 pixels), which results in high scale variations of almost 25 times. To make matters worse, compared to the most popular object detection dataset MS COCO~\cite{ref_article12}, The occlusion between students is very serious. Moreover, The behavior of hand-raising, in different environments, different people and different angles, there are great differences.

In this work, we explore the effectiveness of computer vision techniques in automatically analyzing student behavior patterns in the classroom. Specifically, we focus on hand-raising behavior and have developed a large-scale dataset of labeled images for analysis. The dataset fills a gap in current research on detecting student behavior in classroom teaching scenarios. We have conducted extensive data statistics and benchmark tests to ensure the quality of the dataset, providing reliable training data.

YOLOv7 is one of the best one-stage object detection algorithms currently available, and we attempted to train it on our dataset for better detection results. However, we found that the original version of YOLOv7 still had some room for improvement after training - for example, it would misidentify other actions as raising hands and fail to detect smaller hand-raising actions. Therefore, we incorporated a dynamic sparse attention module called Bi-Level Routing Attention (BRA), which successfully improved detection performance.

\subsubsection{Our main contributions are as follows}:

(1) This paper constructed a publicly available dataset by annotating a large number of images of students raising their hands in the classroom, which supports research on student classroom behavior detection. Compared to existing datasets, SCB-dataset has higher annotation accuracy and more diverse scene samples, filling the data gap in student classroom behavior detection. Additionally, the YOLOv7 object detection algorithm was utilized to train and test the SCB-dataset, achieving satisfactory performance with high practical application value. This work provides a solid foundation and reference for future research in the exploration and application of object detection algorithms in the field of student classroom behavior detection.

(2) This paper proposes an improved model, named YOLOv7-BRA. We added a Bi-Level Route Attention module to the model to give it dynamic query-aware sparsity. Experimental results show that our method successfully improves detection accuracy and reduces false detection rates.

(3) This paper utilizes a fusion of multiple models including YOLOv7 CrowdHuman, SlowFast, DeepSort, and YOLOv7-BRA to detect and obtain student behavior data during classroom sessions, providing essential data for further analysis of student behavior in the classroom. This contribution lays the foundation for future research in the field of student classroom behavior analysis.

\section{Related word}

\subsection{Student classroom behavior dataset}
In recent years, many researchers have adopted computer vision technology to automatically detect students' classroom behaviors, but the lack of open student behavior dataset in the field of education has severely limited the application of video behavior detection in this field. Many researchers have also proposed many unpublished datasets, such as Fu R~\cite{ref_article13} et al, construct a class-room learning behavior dataset named as ActRec-Classroom, which include five categories of s listen, fatigue, hand-rising, sideways and read-write with 5,126 images in total. And R Zheng~\cite{ref_article14} et al, build a large-scale student behavior dastaset from thirty schools, labeling these behaviors using bounding boxes frame-by-frame, which contains 70k hand-raising samples, 20k standing samples, and 3k sleeping samples. and Sun B~\cite{ref_article15} et al,presents a comprehensive dataset that can be employed for recognizing, detecting, and captioning students’ behaviors in a classroom. Author collected videos of 128 classes in different disciplines and in 11 classrooms. However, the above datasets are from real monitoring data and cannot be made public.

\subsection{Students classroom behavior detection}
Mature object detection is used by more and more researchers in student behavior detection, such as YAN Xing-ya~\cite{ref_article4} et al. proposed a classroom behavior recognition method that leverages deep learning. Specifically, they utilized the improved Yolov7 target detection algorithm to generate human detection proposals, and proposed the BetaPose lightweight pose recognition model, which is based on the Mobilenetv3 architecture, to enhance the accuracy of pose recognition in crowded scenarios. And ZHOU Ye~\cite{ref_article16} et al has proposed a method for detecting students' behaviors in class by utilizing the Faster R-CNN detection framework. To overcome the challenges of detecting a wide range of object scales and the imbalance of data categories, the approach incorporates the feature pyramid and prime sample attention mechanisms.

\subsection{Attention mechanisms}
Attention is a crucial mechanism that can be utilized by various deep learning models in different domains and tasks. The beginning of the attention mechanisms we use today is often traced back to their origin in natural language processing~\cite{ref_article17}. The Transformer model proposed in ~\cite{ref_article18} represents a significant milestone in attention research as it demonstrates that the attention mechanism alone can enable the construction of a state-of-the-art model. Recently, sparse attention has gained popularity in the realm of vision transformers due to the remarkable success of the Swin transformer~\cite{ref_article19}. Several works endeavor to make the sparse pattern adaptable to data, including DAT~\cite{ref_article20}, TCFormer~\cite{ref_article21}, and DPT~\cite{ref_article22}. Additionally, BiFormer~\cite{ref_article23} proposes a new dynamic sparse attention approach via bi-level routing to enable a more flexible allocation of computations with content awareness.

\subsection{Student Behavior Detection System}
Various methods and technologies can be used to detect student classroom behavior.  Ngoc Anh B et al~\cite{ref_article24}. developed a computer vision-based application to identify students paying attention in the classroom. Lin et al~\cite{ref_article25}. proposed a student behavior recognition system based on skeleton pose estimation and person detection. Trabelsi et al~\cite{ref_article26}. used machine learning to train models for student behavior recognition, incorporating facial expression recognition for attention detection. Yang~\cite{ref_article27} proposed using YOLOv5, SlowFast and Deep Sort~\cite{ref_article28} for detecting spatiotemporal behavior. Combining different methods and technologies, such as skeleton pose estimation, person detection, and facial expression recognition, and spatiotemporal behavior detection, can improve recognition accuracy and efficiency for student behavior detection.

\section{SCB-dataset}

\begin{figure}
\includegraphics[width=\textwidth]{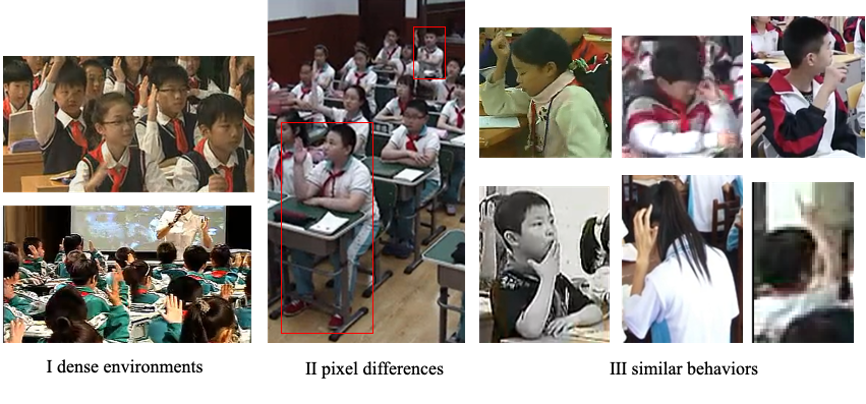}
\caption{Challenges in the student hand-raising behavior dataset include dense environments, similar behaviors, and pixel differences.} \label{fig2}
\end{figure}

\begin{figure}
\includegraphics[width=\textwidth]{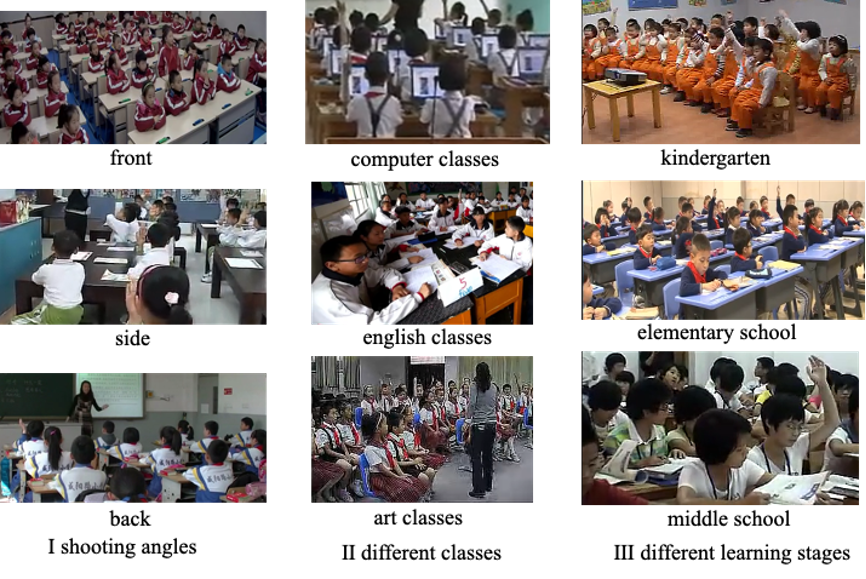}
\caption{Challenges in the student hand-raising behavior dataset include varying shooting angles, class differences, and different learning stages.} \label{fig3}
\end{figure}

Classroom teaching has always played a fundamental role in education. Understanding students' behavior is crucial for comprehending their learning process, personality, and psychological traits. In addition, it is an important factor in evaluating the quality of education. Among different student behaviors, hand-raising behavior is an essential indicator of the quality of classroom participation. However, the lack of publicly available datasets poses a significant challenge for AI research in the field of education.

To address this issue, we have developed a publicly behavior dataset that specifically focuses on hand-raising behavior. Due to the complexity and specificity of educational settings, this dataset presents unique characteristics and challenges that could lead to new opportunities for researchers. The subsequent sections provide detailed information on the dataset's composition and structure.

In reality, people's behavior is often multifaceted and abundant. To capture this complexity, we collected image materials directly from actual classroom recordings available on the bjyhjy and 1s1k websites. By using real-world videos, we ensured that our dataset is representative of actual classroom situations, providing a more realistic and accurate reflection of student behavior.

Classrooms are densely populated environments where multiple subjects engage in different actions simultaneously. For instance, a classroom may have over 100 students present at the same time, as shown in Fig.~\ref{fig2} \uppercase\expandafter{\romannumeral1}. Besides each sitting in various positions, resulting in significant variation in picture sizes in the images, as shown in Fig.~\ref{fig2} \uppercase\expandafter{\romannumeral2}. These conditions create significant challenges for detection tasks.

Detecting hand-raising behavior can be challenging due to the visual similarities it shares with other behavior classes. As shown in Fig.~\ref{fig2} \uppercase\expandafter{\romannumeral3}, we can observe that some action classes exhibit a high degree of visual similarity to hand-raising, which poses significant challenges for detection tasks.

The images in our dataset were captured from different shooting angles, including front, side, and back views, as shown in Fig.~\ref{fig3} \uppercase\expandafter{\romannumeral1}. These angles can significantly impact the visual appearance of students' hand-raising behaviors, further complicating the detection task.

Moreover, the classroom environment and seating arrangement can vary from one course to another, as illustrated in Fig.~\ref{fig3} \uppercase\expandafter{\romannumeral2}, This variability adds another layer of complexity to the detection and recognition of hand-raising behavior.

Additionally, students' hand-raising behaviors can differ significantly at various learning stages, as shown in Fig.~\ref{fig3} \uppercase\expandafter{\romannumeral3}, where we compare kindergarten, elementary school, middle school, and high school behaviors. These differences pose significant challenges for detecting hand-raising behavior across different stages of education.

\begin{table}
\centering
\caption{Statistics of SCB-dataset.}\label{tab1}
\begin{tabular}{|c|c|c|}
\hline
images &  total babels & person/image\\
\hline
4001 &  11248 & 2.81\\
\hline
\end{tabular}
\end{table}

 Table~\ref{tab1} represents the statistical analysis of y based on a dataset of 4001 pictures and 1,248 annotations. On average, each annotation marks 2.81 individuals.

\begin{figure}
\includegraphics[width=\textwidth]{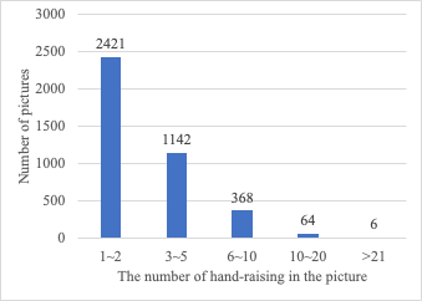}
\caption{Statistical Analysis of the Number of Hand-Raisings in the SCB-dataset.} \label{fig4}
\end{figure}

Fig.~\ref{fig4}  presents statistical data on the number of hand-raising among students in the picture. Specifically, there are 2,421 pictures in which 1~2 students are seen raising their hands, and 1,142 pictures in which 3~5 students are seen raising their hands.

\section{YOLOv7+BRA}

\begin{figure}
\includegraphics[width=\textwidth]{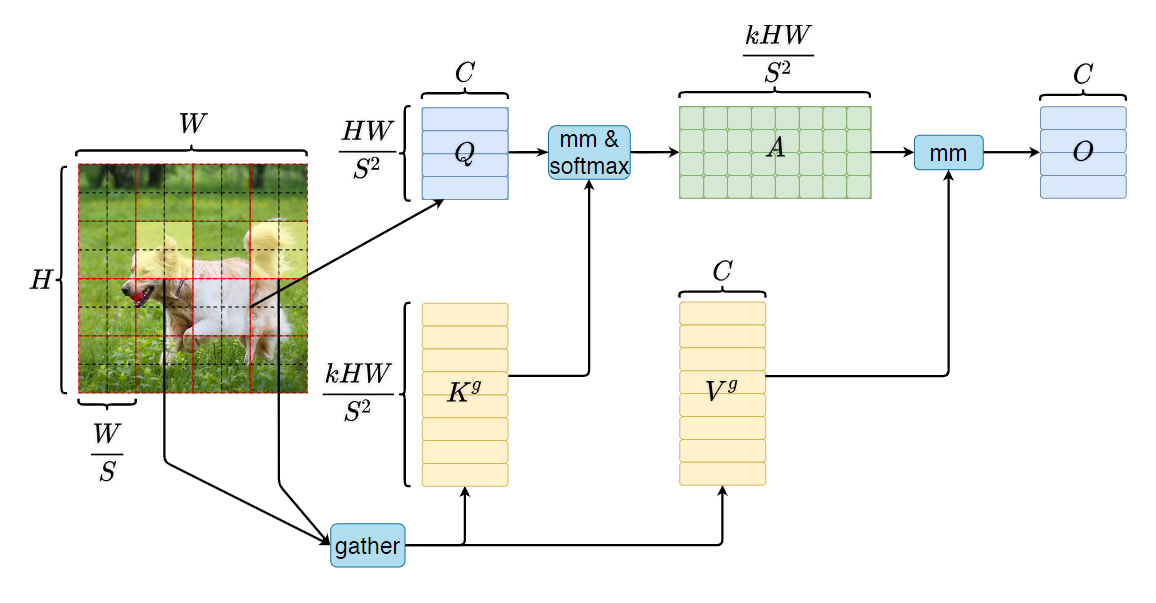}
\caption{Bi-level Routing Attention.} \label{BRA}
\end{figure}

\begin{figure}
\includegraphics[width=\textwidth]{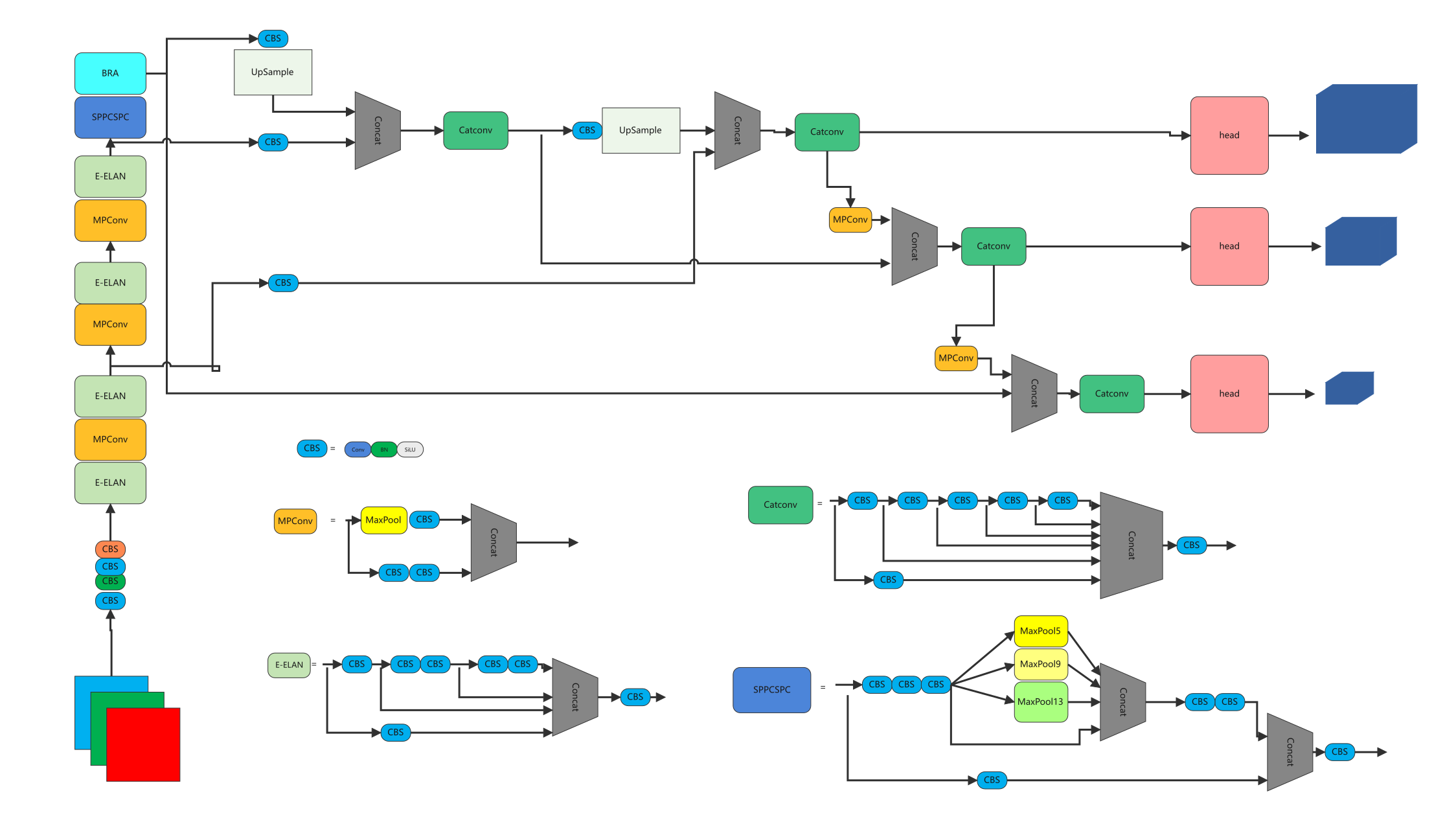}
\caption{the architecture of YOLOv7-BRA.} \label{YOLOv7-BRA2}
\end{figure}
Due to the challenges presented by dense environments, pixel differences and similar behaviors in student classroom behavior monitoring tasks, we have selected YOLOv7 as the fundamental model for our approach after comprehensive consideration of both speed and accuracy. It is a relatively lightweight one-stage object detection algorithm with an inference speed of 6.9ms per image, or a frame rate of over 140fps. This can more or less meet the real-time monitoring needs for student classroom behavior detection.

Generally, one-stage object detection models can be divided into three parts: backbone, neck and head. The purpose of the backbone is to extract and select features, the neck is to fuse features, and the head is to predict results. However, YOLOv7 only retains the backbone and head parts because it proposes an Extended efficient layer aggregation networks (E-ELAN) module to replace various FPNs and PANs commonly used for feature fusion in the neck. Additionally, the Model scaling operation is common in concatenation-based models, which increases the input width of the subsequent transmission layer. Therefore, YOLOv7 proposes the compound scaling up depth and width method.

Despite being one of the best object detection models available, we found that YOLOv7 struggled with handling occlusions and distinguishing similar actions when detecting the SCB dataset. Therefore, we introduced the bi-level routing attention (BRA) module to YOLOv7. BRA is a novel dynamic sparse attention that achieves more flexible computation allocation and content awareness, allowing the model to have dynamic query-aware sparsity. The key to BRA is filtering out most of the irrelevant key-value pairs at a coarse region level, so that only a small portion of routed regions remain. The whole algorithm is summarized with Torch-like pseudo code in Algorithm1

\begin{lstlisting}
# input: features (H, W, C). Assume H==W.# output: features (H, W, C).# S: square root of number of regions.# k: number of regions to attend.
# patchify input (H, W, C) -> (Sˆ2, HW/Sˆ2, C)x = patchify(input, patch_size=H//S)
# linear projection of query, key, valuequery, key, value = linear_qkv(x).chunk(3, dim=-1)
# regional query and key (Sˆ2, C)query_r, key_r = query.mean(dim=1), key.mean(dim=1)
# adjacency matrix for regional graph (Sˆ2, Sˆ2)A_r = mm(query_r, key_r.transpose(-1, -2))
# compute index matrix of routed regions (Sˆ2, K)I_r = topk(A_r, k).index
# gather key-value pairskey_g = gather(key, I_r) # (Sˆ2, kHW/Sˆ2, C)value_g = gather(value, I_r) # (Sˆ2, kHW/Sˆ2, C)
# token-to-token attentionA = bmm(query, key_g.transpose(-2, -1))A = softmax(A, dim=-1)
output = bmm(A, value_g) + dwconv(value)
# recover to (H, W, C) shapeoutput = unpatchify(output, patch_size=H//S)
\end{lstlisting}
bmm: batch matrix multiplication; mm: matrix multiplication. dwconv: depthwise convolution

The process of BRA can be easily divided into three steps: firstly, assuming we input a feature map, we divide it into several regions, and obtain query, key, and value through linear mapping. Secondly, we use the adjacency matrix to build a directed graph to find the participating relationship corresponding to different key-value pairs, which can be understood as the regions that each given region should participate in. Finally, with the routing index matrix from region to region, we can apply fine-grained token-to-token attention.

The structure of our modified YOLOv7-BRA (YOLOv7 with Bi-level Routing Attention) model is shown in the Fig.~\ref{YOLOv7-BRA2}. We place the BRA module in the final part of the backbone. When introducing the BRA module, we considered placing it in three different positions: (1) replacing all convolutions with convolutions that include BRA; (2) placing BRA in the head section; and (3) placing BRA in the backbone section. If we choose method (1), it will result in a very large model that is difficult to train and affects inference speed. As for whether to place BRA in the head or in the backbone, considering that the role of the attention mechanism is to make the model only focus on specific areas of the image rather than the entire image, we believe this is part of feature extraction. Therefore, we chose to place it in the backbone section.

\section{Student Classroom Behavior Detection System}

\begin{figure}
\includegraphics[width=\textwidth]{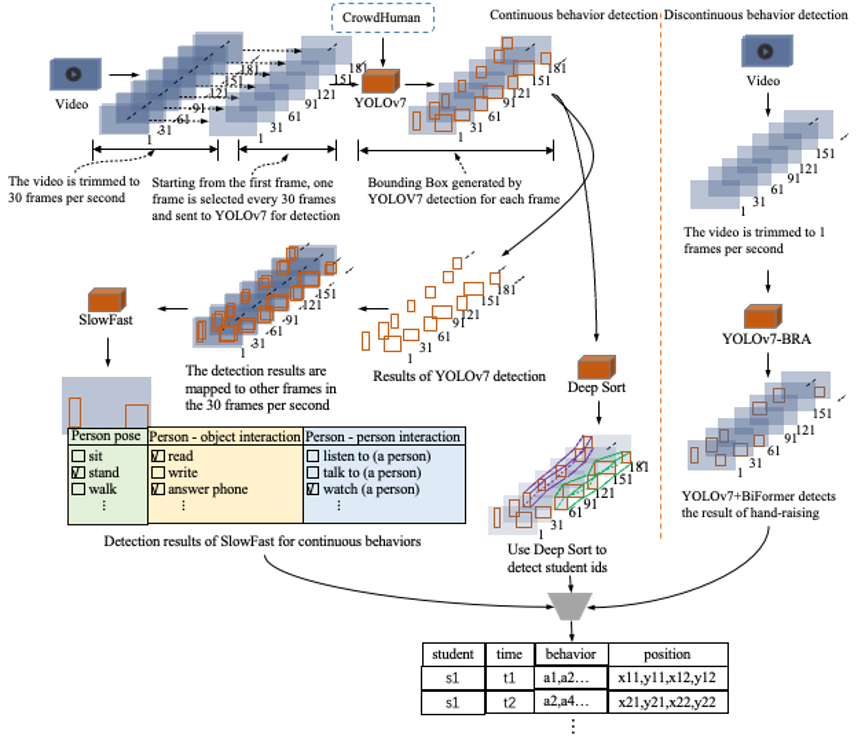}
\caption{Process of Student Classroom Behavior Detection System.} \label{fig5}
\end{figure}

Fig.~\ref{fig5} shows the detailed process of the student classroom behavior detection system. The figure is divided into three main parts: detection of continuous student behaviors, detection of non-continuous student behaviors, and the fusion of behavior detection results with student IDs.

For the detection of continuous student behavior, the video is first sampled at 30 frames per second, and YOLOv7 is used to perform detection every 30 frames, with weights trained on the CrowdHuman dataset to adapt to dense classroom scenes. The detection results are then sent to both Deep Sort and SlowFast, with SlowFast mapping the results to other frames within the same second. SlowFast can detect continuous behaviors, which are mainly classified into person pose, person-object interaction, and person-person interaction, such as sit, stand, read, write, talk, etc.

For the detection of non-continuous student behavior, the video is first sampled at 1 frame per second, and YOLOv7-BRA is used to detect raising hands.

For the fusion of behavior detection results with student IDs, the continuous and non-continuous behavior detection results are merged with the student IDs detected by Deep Sort, resulting in student IDs, time, behavior, and location information that are essential for student classroom behavior analysis.

\begin{figure}
\includegraphics[width=\textwidth]{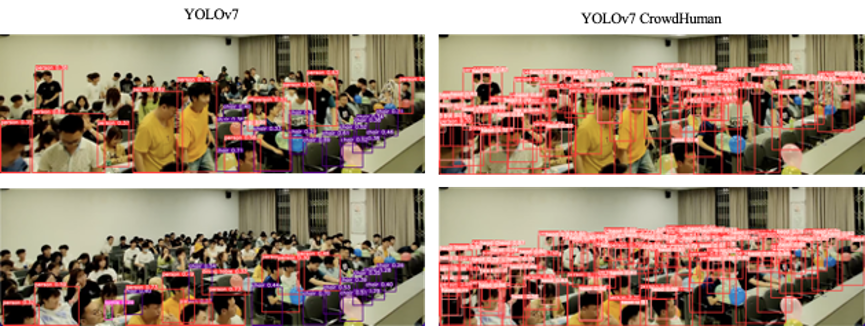}
\caption{Comparison of Detection Results between YOLOv7 and YOLOv7 CrowdHuman.} \label{fig6}
\end{figure}

Additionally, in the continuous behavior detection process, we used YOLOv7 as the student detection network to detect the coordinates of the students in the video frames. However, we found that YOLOv7 had poor detection results in classroom scenarios, with many misses for students in the back rows. Therefore, we used the YOLOv7 weights trained on CrowdHuman, a dataset designed for dense scenarios, and found that these weights were also suitable for the classroom scene. The detection results of YOLOv7 and YOLOv7 trained on CrowdHuman are compared in Fig.~\ref{fig6}.

As shown in Fig.~\ref{fig6}, the left side displays the detection results of YOLOv7, while the right side displays the detection results of YOLOv7 trained on CrowdHuman. It is evident that the detection results of YOLOv7 trained on CrowdHuman are significantly better than those of YOLOv7.

\begin{figure}
\includegraphics[width=\textwidth]{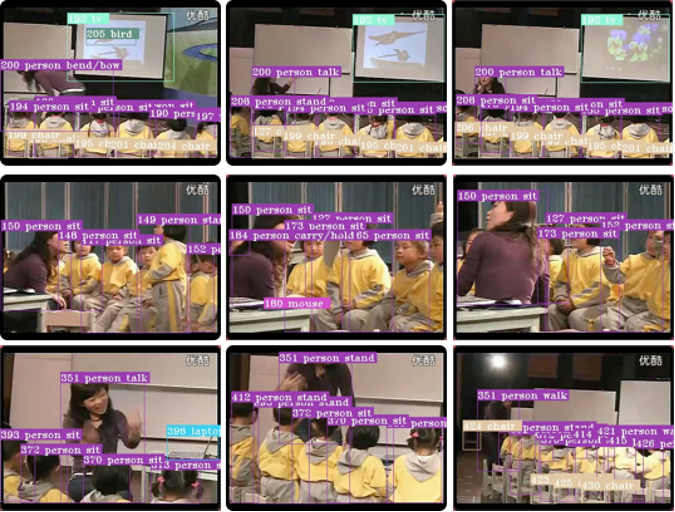}
\caption{Detection and tracking results based on YOLOv7 CrowdHuman, SlowFast, and Deep Sort.} \label{fig7}
\end{figure}

As shown in Fig.~\ref{fig7} our detection results based on YOLOv7 CrowdHuman, SlowFast, and Deep Sort demonstrate that ID tracking for both students and teachers is relatively stable within the same angle of video frames. Additionally, the continuity of their behaviors is also accurate. However, it can be observed that the ID of students and teachers changes when the angle of video frames changes. This indicates that further improvements are needed in our future work.

\section{Experiment and Analysis}
\subsection{Experimental Environment and Dataset}
We conducted our experiments on an NVIDIA GeForce RTX 2080 Ti GPU with 11 GB of video memory, running Ubuntu 20.04.2 as the operating system. The code was implemented in Python 3.8 and we used PyTorch version 1.8.1 with CUDA version 10.1 for model training.

The dataset used in our experiments is SCB-dataset, which we split into training, validation sets with a ratio of 4:1.

\subsection{Experimental content}
To validate the effectiveness of the proposed YOLOv7-BRA, experiments were conducted from the following perspectives: Comparison of the detection accuracy and performance between YOLOv7-BRA model and various models of YOLOv7, as well as YOLOv5 model.

\subsection{Model Training}
The training process consists of three parts. The first part involves training various network architectures of YOLOv7, followed by training the YOLOv5m network architecture in the second part, and finally training the YOLOv7-BRA network architecture in the third part.

To train the model, set the epoch to 150, batch size to 8, and image size to 640x640, and we use a pre-trained model for the training.

\subsection{Evaluation Metrics}
In order to objectively analyze the experimental results, we use the Mean Average Precision (mAP) as an evaluation index, with an IOU of 0.5. The formulas  is as follows:
\begin{equation}
Recall=TP/(TP+FN)
\label{eq:1} 
\end{equation}

\begin{equation}
Precision=TP/(TP+FP)
\label{eq:2} 
\end{equation}

\begin{equation}
mAP = \int_0^1 \left( P(R) \right) dR
\label{eq:3} 
\end{equation}

Formula (\ref{eq:1}) Recall, denoted as R, represents the recall rate. Formula (\ref{eq:2}): Precision, denoted as P, represents the precision. TP (True Positive) represents the number of positive samples that are correctly identified, FN (False Negative) represents the number of positive samples that are incorrectly identified as negative, FP (False Positive) represents the number of negative samples that are incorrectly identified as positive, TP + FN represents the total number of positive samples, TP + FP represents the total number of samples that are identified as positive, TP and FP are determined based on the IOU (Intersection Over Union) threshold. The formula for calculating IOU is as follows:

\begin{equation}
IOU(A,B)=|(A\cap B)/(A\cup B)|
\label{eq:4} 
\end{equation}

In which A represents the ground truth box, and B represents the box predicted based on anchors and detected by the model.

\subsection{Experimental Results and Analysis}

For our training, we utilized various network structures from YOLOv7 such as YOLOv7-tiny, YOLOv7, YOLOv7-X, YOLOv7-W6, YOLOv7-E6  and, we employed YOLOv7-BRA and YOLOv5m network structures.  The results of our experiments are outlined in Figure 8, with precision denoted as "p" and recall denoted as "R".

\begin{figure}
\includegraphics[width=\textwidth]{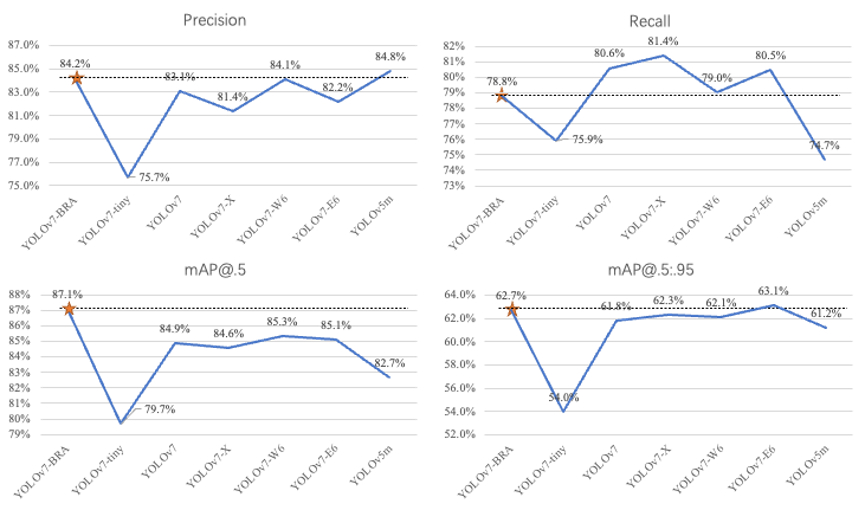}
\caption{Evaluation of hand-raising on SCB-dataset.} \label{fig8}
\end{figure}

From Fig.~\ref{fig8}, it can be seen that YOLOv-BRA has higher Precision results than the YOLOv7 series models.   Looking at the mAP@0.5 results, the YOLOv-BRA model outperforms YOLOv7 series models and yolov5m model, with a difference of 2.2\% over the second-place model.   In terms of mAP@0.9, the YOLOv-BRA model outperforms all other YOLO series models except for YOLOv7-E6, which has a much more complex network structure and requires more training time, YOLOv-BRA also outperforms the yolov5m model in this regard.

\begin{figure}
\includegraphics[width=\textwidth]{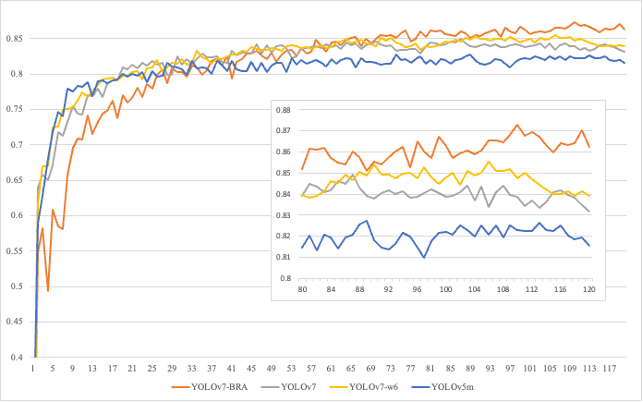}
\caption{Comparison of mAP@0.5 Results for Each Network's Training Iterations.} \label{fig9}
\end{figure}

Fig.~\ref{fig8} shows the results of mAP@0.5 for YOLOv7-BRA, YOLOv7, YOLOv7-w6, and YOLOv5m during the training iterations.  It can be observed from the figure that the accuracy of YOLOv7-BRA is lower than that of the other networks in the first 30 iterations.  However, after 70 iterations, the accuracy of YOLOv7-BRA surpasses that of the other networks.

\begin{figure}[H]
\includegraphics[width=\textwidth]{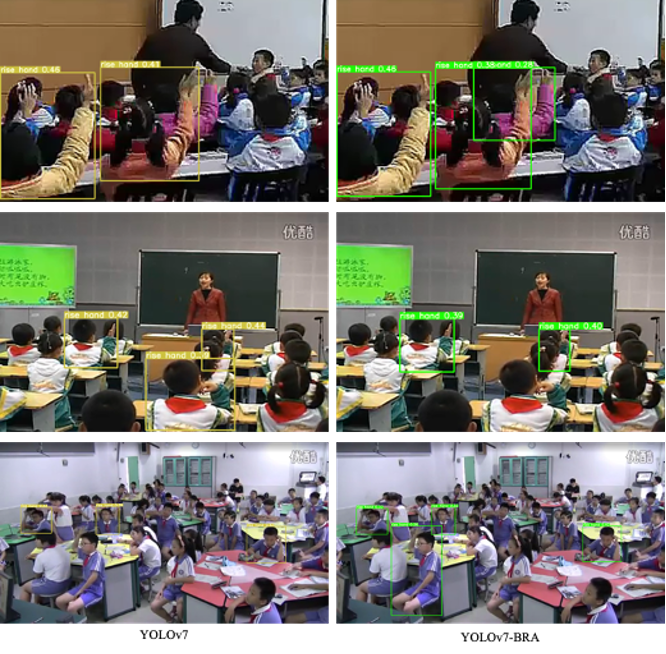}
\caption{Comparison of detection results between YOLOv7 and YOLOv7-BRA models.} \label{fig10}
\end{figure}

As shown in Fig.~\ref{fig10}, a comparison is made between the detection results of YOLOv7 and YOLOv7-BRA.  In the comparison image of the first row, it can be observed that YOLOv7-BRA detected situations where the arms of the front-row students were raised, despite severe obstruction.  From the comparison image of the second row, it can be seen that YOLOv7-BRA did not make the mistake of mistaking a student resting their hand on their head for raising their hand.  Finally, in the comparison image of the third row, it can be seen that YOLOv7-BRA was able to detect students raising their hands even in the presence of clutter in the background.

\section{Conclusion}
This paper emphasizes the importance of accurately detecting student behavior in classroom videos and proposes the Student Classroom Behavior Detection system based on YOLOv7-BRA.  The research identified eight behavior patterns, constructed a dataset of over 4,000 images with 11,248 labels, and used the bi-level routing attention module to improve detection accuracy.  The fusion of multiple models, including YOLOv7-CrowdHuman, SlowFast, DeepSort, and YOLOv7-BRA, effectively obtained student behavior data during classroom sessions.  The contributions of this paper including the SCB-dataset and improved YOLOv7-BRA model provide a solid foundation for future research and development in the field of student behavior detection, ultimately benefiting students' educational outcomes.  Future work should focus on increasing the quantity and category of student behavior datasets, using diverse networks to provide reliable data references, and addressing limitations such as DeepSort's performance changes in cases of video angle changes. Overall, this research has a significant role in advancing educational technology and improving teaching effectiveness.

%
%
%
%

\end{document}